\documentclass[a4paper]{article}
\usepackage{dx}
\usepackage[T1]{fontenc}
\usepackage{ae,aecompl}
\usepackage{amsfonts}
\usepackage{amsmath}
\usepackage{amsthm}
\usepackage{enumitem}
\usepackage{graphicx}
\usepackage{times}

\usepackage{comment}
\usepackage{varwidth}
\usepackage{balance}

\newtheorem{definition}{Definition}

\newcommand{\sd}{\textsc{sd}}
\newcommand{\comps}{\textsc{comps}}
\newcommand{\obs}{\textsc{obs}}
\newcommand{\meas}{\textsc{meas}}
\newcommand{\md}{\Delta}
\newcommand{\mc}{\mathcal{C}}
\newcommand{\OK}{\textsc{ok}}
\newcommand{\ok}{\mathit{ok}}
\newcommand{\jobs}{\mathit{Jobs}}
\newcommand{\fw}{\mathit{Framework}}
\newcommand{\cost}{\mathit{Cost}}
\newcommand{\Ops}{\mathit{Ops}}
\newcommand{\type}{\mathit{type}}
\newcommand{\length}{\mathit{length}}
\newcommand{\start}{\mathit{start}}
\newcommand{\arrival}{\mathit{arrival}}
\newcommand{\beh}{\mathit{beh}}
\newcommand{\mb}{\mathcal{B}}
\newcommand{\machines}{\mathit{Ma}}

\newtheorem{thm}{Theorem}

\newcounter{examplecounter}
\newenvironment{example}{
	\refstepcounter{examplecounter}%
	
	\vspace{4pt}
	\noindent\textbf{Example \arabic{examplecounter}}%
	\quad
}{
	
	\vspace{4pt}
}

\newcounter{scenariocounter}
\newenvironment{scenario}{
	\refstepcounter{scenariocounter}%
	
	\vspace{3pt}
	\noindent\textbf{Scenario \arabic{scenariocounter}}%
	\quad
}{
	
	\vspace{3pt}
}

\title{The Scheduling Job-Set Optimization Problem: \\ A Model-based Diagnosis Approach\thanks{This work was accepted for presentation at the \emph{31st International Workshop on Principles of Diagnosis (DX-2020)}. A version of this paper presenting a more sophisticated algorithm as well as an extended evaluation is published in the \emph{Proceedings of the International Conference on Principles of Knowledge Representation and Reasoning 2021 (KR-2021)} \protect\cite{rodler2021kr}.}}

\author%
{%
Patrick Rodler \and 
Erich Teppan\thanks{Authors in alphabetical order.}\\
Universit\"at Klagenfurt \\
e-mail: firstname.lastname@aau.at\\
}

\begin{document}

\maketitle

\begin{abstract}
A common issue for companies is that the volume of product orders may at times exceed the production capacity. We formally introduce two novel problems dealing with the question which 
orders to discard or postpone in order to meet certain (timeliness) goals, and try to approach them by means of model-based diagnosis. In thorough analyses, we identify many similarities of the introduced problems to diagnosis problems, but also reveal crucial idiosyncracies and outline ways to handle or leverage them. Finally, a proof-of-con-cept evaluation on industrial-scale problem instances from a well-known scheduling benchmark suite demonstrates that one of the 
two formalized problems can be well attacked by out-of-the-box model-based diagnosis tools. 
\end{abstract}

\section{Introduction}
\label{sec:intro}
The scheduling of jobs~\cite{Blazewicz:2007} is an important task in almost all production systems and there exist various objectives to be optimized like, e.g., completion time (time to finish all products), or tardiness (lateness of products).
A job is here associated with the production of one product item.
There is a long history of scheduling research and many different approaches have been used to solve scheduling problems. Aiming at optimal solutions with exact methods, 
e.g., constraint-based approaches (e.g.\ \cite{cspsched}), branch and bound (e.g.\ \cite{BRUCKER1994107}), branch and cut (e.g.\ \cite{Stecco}) or mixed integer programming (e.g.\ \cite{KU2016165}) have been successfully employed. For near-optimum solutions, current state-of-the-art approaches are based on tabu and large neighborhood search (e.g.\ \cite{BOZEJKO2017512,lns}), simulated annealing or genetic algorithms (e.g.\ \cite{Sadegheih2006147}).

Despite these extensive research efforts on how to calculate optimal or near optimal schedules for different types of job scheduling problems, the challenge of reconciling business objectives with customer needs in cases where even optimal schedules do not satisfy 
given time constraints
has not been addressed so far. Take as an important example the following. Many 
companies face (at least during certain periods) the problem that the amount of products ordered by customers exceeds the production capacity for a certain planning horizon. Hence, there are more jobs than can be accomplished. Depending on the type of product and the market situation, different scenarios are possible, e.g.:

\begin{scenario} \hspace{-1em}\emph{(Low customer loyalty / many competitors)} Examples can be found in low-tech products where the production know-how is possessed by many market competitors. Due to low customer loyalty in this case, the risk of losing orders that cannot be produced within certain deadlines is relatively high. Consequently, a subset of product orders (i.e., a subset of jobs) that maximizes a specified company target like revenue should be prioritized.
\end{scenario}
\vspace{-3pt}
\begin{scenario} \hspace{-1em}\emph{(High customer loyalty / few competitors)} Examples for that scenario can be found in specialized high-tech products, e.g., life-critical airbag chips for cars. Normally, customer loyalty is high in such a case as there are not many companies possessing the needed know-how and customer relationships are based on trust. As a consequence, the chance to lose the product orders if not able to produce within the current planning horizon is lower. Nevertheless, a job should be postponed only for good reasons, as this might displease a customer and become a problem in the long run. Hence, the set of postponed jobs should be subset-minimal.
\end{scenario}

Based on these scenarios, we identify two problems that appear to be strongly related to model-based diagnosis problems \cite{Reiter87}. The first one, the \emph{Job-set Optimization Problem} (\emph{JOP}, associated with Scenario~1) is similar to the problem of computing a minimum-cardinality or most-preferred diagnosis. The second one, the \emph{Job-set Maximization Problem} (\emph{JMP}, associated with Scenario~2) resembles the problem of computing a minimal diagnosis.

The contribution of this paper is threefold: \emph{(a)}~In Sec.~\ref{sec:prob_def} 
we formally define JOP and JMP based on a generic characterization of job scheduling problems, and exemplify a concrete instantiation of these by means of the job shop scheduling problem. 
\emph{(b)}~In Sec.~\ref{sec:JOP_vs_MBD} we thoroughly discuss \emph{(i)}~similarities and differences between model-based diagnosis and JMP/JOP, \emph{(ii)}~how diagnostic concepts can be profitably used to approach JMP/JOP, and \emph{(iii)}~ways of adapting diagnostic approaches in order to leverage the idiosyncracies of JMP/JOP. 
\emph{(c)}~After the discussion of related work in Sec.~\ref{sec:related_work}, we provide a proof of concept in Sec.~\ref{sec:eval} showing that methods from model-based diagnosis can be successfully applied to JMP, and we address ways of approaching JOP based on our observations made. Concluding remarks and pointers to future work are given in Sec~\ref{sec:conclusion}. 

\section{Problem Definition}
\label{sec:prob_def}
We first provide a meta-level definition which characterizes a general framework 
to capture different manifestations of the job scheduling problem, such as (flexible) job shop, flow shop or open shop problems (see \cite{Blazewicz:2007} for an overview). 

\begin{definition}
\label{def:job_scheduling_problem}
Generally, any \emph{job scheduling problem} $P = \langle\jobs, \fw, \cost\rangle$ comprises:

\begin{enumerate}[noitemsep]
\item A set of $\jobs = \{j_1, \dots, j_p\}$, where every $j \in \jobs$ has a predefined arrival time $\arrival_j$ and a set of operations $\Ops_j = \{op_1, \dots, op_{l_j}\}$. 
Each operation $op \in \Ops_j$ has a predefined operation type $\type_{op}$ and operation length $\length_{op} \in \mathbb{N}$ as part of the input and a start time $\start_{op} \in \mathbb{N}_0$ that is to be assigned.
\begin{itemize}
\item We call a complete assignment of start times a \emph{schedule}.
\end{itemize}
\item A formal description of the $\fw$ in which the jobs are to be performed. Depending on the problem variant at hand, such a framework description typically includes constraints about capabilities of resources/machines that perform the job operations and precedence constraints concerning jobs and operations.
\begin{itemize}
\item We call a $schedule$ 
not violating framework constraints a \emph{consistent schedule}.
\end{itemize}
\item A function $\cost: S \rightarrow \mathbb{N}$, that assigns a cost to each consistent schedule $S$.
\begin{itemize}
\item An \emph{optimal solution/schedule}
for a scheduling problem $P$ is a consistent schedule $S$ such that $\cost(S) \rightarrow \min$.
\end{itemize}
\end{enumerate}
\label{def1}
\end{definition}


We will use the NP-complete \emph{job shop scheduling problem} \cite{Blazewicz:2007} as a prototypical instance of job scheduling throughout this work:

\begin{example} \hspace{-1em}\emph{(Definition of the Job Shop Scheduling Problem)}\quad\label{ex:job_shop_scheduling_with_completion_time_defined_using_DEF1} An instance of the job shop scheduling problem with completion-time optimization can be expressed in the form of Def. \ref{def1} as $P = \langle\jobs, \fw, \cost\rangle$:

\begin{enumerate}[noitemsep]
    \item In the static version of the problem all jobs exist from the very beginning. Thus, all job arrival times are set to zero, i.e., $\arrival_j=0$ for all $j \in \jobs$.
    \item The $\fw$ defines the following rules:
    \begin{itemize}
        \item The set of operations $\Ops_j$ of a job $j$ builds a sequence. An operation $op_x \in \Ops_j$ must end before the succeeding operation $op_{x+1} \in \Ops_j$ can start, i.e., $\start_{op_{x+1}} \geq \start_{op_x} + \length_{op_x}$.
        \item There is a set of machines $\machines = \{m_1, \dots, m_q\}$ each of which performs operations one-by-one (non-preemptive). For each operation $op$ there is exactly one machine $machine_{op} \in \machines$ where $op$ must be performed.
    \end{itemize}
    \item The cost of a schedule is its completion time, i.e., the time when the latest operation is finished. More formally: Given a job scheduling problem $P$ and a corresponding schedule $S$, $\cost(S) = max(\start_{op_j} + \length_{op_j})$, with $op_j \in \Ops_j, j \in \jobs$.
\end{enumerate}
\end{example}
A concrete instance of the job shop scheduling problem is discussed next.

\begin{figure*}[t]
\centering
\includegraphics[width=14cm]{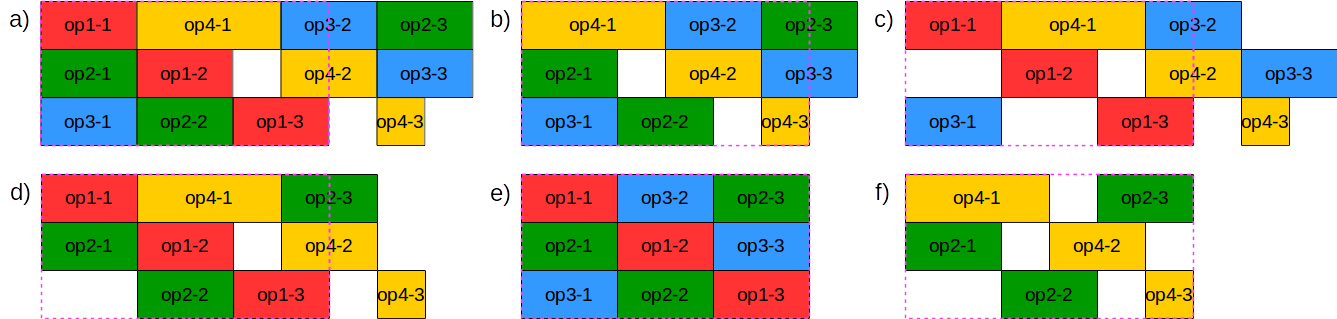}
\caption{\small Example~\ref{ex:job_shop_scheduling_illustration_of_JMP+JOP}. a) Optimal solution for all jobs; b) Optimal solution after removal of job 1; c) Optimal solution after removal of job 2; d) Optimal solution after removal of job 3; e) Optimal solution after removal of job 4; f) Optimal solution after removal of jobs 1 and 3.}
\label{fig:exAll}
\end{figure*}

\begin{example} \hspace{-1em}\emph{(Instance of Job Shop Scheduling Problem)}\quad \label{ex:job_shop_scheduling_example} 
Consider the following simple job shop scheduling problem comprising four jobs consisting of three operations each, which have to be processed by three machines: 

\begin{center}
\begin{varwidth}{\linewidth}
{\footnotesize{
\begin{verbatim}
Job 1:  op1-1 (type=machine1 | length=2)
        op1-2 (type=machine2 | length=2)
        op1-3 (type=machine3 | length=2)

Job 2:  op2-1 (type=machine2 | length=2)
        op2-2 (type=machine3 | length=2)
        op2-3 (type=machine1 | length=2)
\end{verbatim}
}}
\end{varwidth}
\end{center}

\begin{center}
\begin{varwidth}{\linewidth}
{\footnotesize{
\begin{verbatim}
Job 3:  op3-1 (type=machine3 | length=2)
        op3-2 (type=machine1 | length=2)
        op3-3 (type=machine2 | length=2)

Job 4:  op4-1 (type=machine1 | length=3)
        op4-2 (type=machine2 | length=2)
        op4-3 (type=machine3 | length=1)
\end{verbatim}
}}
\end{varwidth}
\end{center}

An optimal solution for this problem has a completion time of 9 (see Fig.~\ref{fig:exAll}a).
\end{example}

We next define the \emph{job-set maximization / optimization problem} which is of relevance if not all given jobs in $\jobs$ can be processed due to cost (time, resources, etc.) constraints. 
For instance, 
assume a company that receives more orders than the factory can process until a given deadline. The goal of the company is then to take only a subset of all orders which can be finished in time and which maximizes the company's utility (revenue, etc.). 
\begin{definition}\label{def:JOP_problem}
Given a job scheduling problem $P = \langle\jobs, \fw, \cost\rangle$ and a constant $\kappa \in \mathbb{R^+}$, the \emph{job-set maximization problem (JMP)} $\langle P, \kappa\rangle$ is to find a subset-minimal job set $\Delta \subseteq \jobs$ such that for some consistent schedule $S'$ for $P' = \langle\jobs\setminus\Delta, \fw, \cost\rangle$ it holds that $\cost(S')\leq\kappa$.

If each $j \in \jobs$ is associated with a utility $u_j \in \mathbb{N}$, then the \emph{job-set optimization problem (JOP)} is to find a solution $\Delta^* \subseteq \jobs$ for JMP such that the sum of utilities of jobs in $\jobs \setminus \Delta^*$ is maximal, i.e., $\sum_{j \in \jobs\setminus\Delta^*} u_j \rightarrow \max$. 
\end{definition}
JMP can be seen as a computation problem (\emph{any} subset-minimal solution), and JOP as the associated optimization problem (\emph{best} subset-minimal solution). 
\begin{thm}[Complexity of JMP/JOP] \label{thm1} JMP is NP-easy (and not NP-hard). JOP is NP-hard (and not NP-easy).
\end{thm}

In the following we describe a concrete instance of a job shop scheduling problem and use it to explicate Def.~\ref{def:JOP_problem}. 

\begin{example} \hspace{-1em}\emph{(Illustration of JMP and JOP)}\quad \label{ex:job_shop_scheduling_illustration_of_JMP+JOP}
Reconsider the 
scheduling problem instance given in Example~\ref{ex:job_shop_scheduling_example}. If there is only restricted time of $\kappa = 6$ available, the JMP is to find out which subset-minimal set $\Delta$ of jobs to discard, or, equivalently, which superset-maximal set $\jobs\setminus\Delta$ of jobs can be accomplished within the limited time period $\kappa$. There are two possible solutions to JMP: $\Delta_1 = \{\text{job 4}\}$ (cf.\ Fig.~\ref{fig:exAll}e) and $\Delta_2 = \{\text{job 1, job 3}\}$ (cf.\ Fig.~\ref{fig:exAll}f). Since no utility of jobs is considered in JMP, both solutions are of equal value. 

Now, let us turn our attention to the JOP. When assuming that all jobs have equal utility (e.g., the product corresponding to each job has the same processing cost and the company sells all products with the same revenue), then we search for the $\Delta$ of minimum cardinality, i.e., for the cardinality-maximal set of products whose production can be scheduled within $\kappa$ time units. A systematic search approach would first test all $\Delta$'s of cardinality one. Fig.~\ref{fig:exAll}b--e illustrate the optimal schedules resulting from the four trials removing a single job each. We can see that the removal of job 1 / 2 / 3 allows a completion time of 7 / 9 / 8, and that the only minimum-cardinality solution is $\Delta_1 = \{\text{job 4}\}$.

Finally, let the utility (e.g., product revenue) of jobs 1 to 4 be $\langle u_1,u_2,u_3,u_4 \rangle = \langle 2,3,1,4\rangle$. Then the only solution to JOP is $\Delta_2 = \{\text{job 1, job 3}\}$ since the sum of utilities of jobs in $\jobs\setminus \Delta_2 = \{\text{job 2, job 4}\}$ is $3+4 = 7$. Note, the second possible solution, $\Delta_1$, has a utility of only $2+3+1 = 6$.   
%
\end{example}

\section{Model-Based Diagnosis and JMP/JOP}
\label{sec:JOP_vs_MBD}
Since we want to apply notions and techniques from model-based diagnosis to solve a non-diagnosis problem, it is important to detect potential idiosyncracies resulting from that domain shift which require special attention and handling.  
Thus, we analyze in this section the commonalities and differences between concepts used in model-based diagnosis and their equivalents in job-set maximization / optimization. 

\subsection{Model-Based Diagnosis (MBD)}
Model-based diagnosis \cite{Reiter87,dekleer1987} assumes a system consisting of a set of \emph{components} $\comps$.
For this system, the availability of a formal (logical) \emph{system description} $\sd$ is assumed, where the \emph{normal behavior} $\beh(c)$ of all system components $c \in \comps$ is specified by means of a sentence $\ok(c) \to \beh(c)$. Moreover, $\sd$ includes, e.g., the system structure, general knowledge about the system domain, etc.\ and is formulated using some monotonic knowledge representation language. We denote the assumption that components $X \subseteq \comps$ are working normally, and components $\comps\setminus X$ are at fault, i.e., $\{\ok(c)|c\in X\}\cup\{\lnot\ok(c)|c\in\comps\setminus X\}$, by $\OK(X)$.
\emph{Observations} $\obs$ and/or \emph{measurements} $\meas$ of the real system behavior, encoded as logical statements, are then compared (\emph{consistency checking}) with predictions made by means of theorem provers based on the system description and the assumption that all components are working properly.  
If discrepancies are found,
i.e., $\sd \cup \OK(\comps) \cup \obs \cup \meas \models \bot$, the assumption that all components are fault-free needs to be retracted. The goal is then to find a \emph{diagnosis}, which is a set of components $\md \subseteq \comps$ whose abnormality explains the observed misbehavior, i.e., $\sd \cup \OK(\comps\setminus\md) \cup \obs \cup \meas \not\models \bot$. Useful for diagnosis computation is the notion of a \emph{conflict}, which is a set of components $\mc \subseteq \comps$ of which at least one component must be faulty, i.e., $\sd \cup \OK(\mc) \cup \obs \cup \meas \models \bot$. 
A diagnosis (conflict) is said to be \emph{minimal} iff no proper subset of this diagnosis (conflict) is a diagnosis (conflict); a diagnosis $\md$ is a minimum-cardinality diagnosis iff there is no diagnosis $\md'$ with $|\md'| < |\md|$. The link between diagnoses and conflicts is 
then 
that each (minimal) diagnosis is a (minimal) hitting set of all minimal conflicts \cite{Reiter87}. 
We call $\langle \sd,$ $\comps,\obs,\meas \rangle$ a \emph{diagnosis problem instance (DPI)}. 

Given a DPI, the task is usually to compute one diagnosis or a set of diagnoses. To avoid combinatorial explosion when considering all diagnoses, the focus is normally laid on minimal diagnoses only. If the specification of a DPI is accompanied with some meta-information that allows to extract a failure probability for each component in $\comps$, then this information can be used to specify the \emph{probability of diagnoses} (of being the true diagnosis that pinpoints the abnormal components). The diagnosis computation problem can thereby be restated as a \emph{diagnosis optimization problem}, where minimal diagnoses are to be found in descending order of probability. In the important special case where all components have an equal failure probability (of less than $0.5$ \cite{Rodler2015phd}), the most probable solutions are exactly the minimum-cardinality diagnoses.

\subsection{MBD concepts in JMP/JOP}
We next discuss the effect of carrying over these concepts to the JMP/JOP domain and their interpretation in this domain: 

\noindent\textbf{Components and (Ab)Normality:} 
The components in case of JMP/JOP are the jobs $j \in \jobs$. However, unlike in MBD, there is no notion of "normal behavior" for jobs. That is, $\beh(j)$ is not defined and thus sentences of the form $\ok(j) \to \beh(j)$ are obsolete and do not need to be included in the system description. Rather, a mechanism is needed to (in)activate jobs. Similarly as done in 
KB debugging \cite{rodler17dx_reducing}, we simply (remove) add a particular job $j$ (from) to $\jobs$ to accomplish this (in)activation. 
At the end of the day, however, the assumption $\lnot \ok(c)$ in MBD is similar to a removal of $j$ from $\jobs$ in JMP/JOP, since the former suspends some constraint $\beh(c)$ (involving that some implications might be no longer derivable from $\sd$) and the latter revokes the requirement that $j$ has to appear in the schedule (involving that schedule's cost might have decreased). 
    %
    
\noindent\textbf{System Description:} Basically, there is no \emph{explicit} system description in JMP/JOP like we know from classic MBD applications. E.g., structural relations between components, such as locality or successor relationships, or the components' input-output behavior, which are aspects readable from system descriptions of, e.g., circuits or software programs, are not specified in case of JMP/JOP. Still, there is some form of system description in JOP as well, which is essentially represented by the $\fw$ (cf.\ Def.~\ref{def:job_scheduling_problem}) of the underlying job scheduling problem. In some sense, $\fw$ can be seen as an \emph{implicit} system description since the relationships between the components ($\jobs$) in this model become fully explicit only after a concrete consistent schedule has been built for $\fw$, and might differ between different consistent schedules.  
    
    With that said, the core intuition inherent in MBD of comparing a real artifact with its formal description while assuming all components normal, 
    and finding components whose abnormality explains observed discrepancies, does not apply to JMP/JOP. For there is no real artifact in JMP/JOP. The equivalent to it is an (unknown) consistent schedule that has cost less than the given $\kappa$. This perspective allows us to recognize the original intuition from MBD also in JMP/JOP---detect which components (jobs) "constitute the difference" between the "real artifact" (desired schedule cheaper than $\kappa$) and the "described system under normality assumption" (the task of scheduling all jobs $\jobs$ while meeting the requirements specified by $\fw$). 
    %
 
 \noindent\textbf{Consistency Checking:} The consistency check is one pivotal difference between classic MBD and JMP/JOP. In MBD, it is a logical consistency check realized by a theorem prover in order to test if certain assumptions $\OK(X)$ for $X \in \comps$ are consistent with the system description ($\sd$), the observations ($\obs$) and the measurements ($\meas$). In JMP/JOP, the equivalent to this consistency check is a test "if there is some consistent schedule $S$ for a given JMP/JOP instance $\langle P,\kappa\rangle$ such that $\cost(S) \leq \kappa$." This test is performed by a scheduling-problem solver,
 i.e., an algorithm for solving the 
 job scheduling problem (cf.\ Def.~\ref{def:job_scheduling_problem}).
 %
 We make two important observations: \emph{(i)}~\emph{The definition of the consistency check in JOP is dependent on the continuous numeric parameter} $\kappa$ (the maximal allowed schedule cost, cf.\ Def.~\ref{def:JOP_problem}). \emph{(ii)} Since part of the considered JMP/JOP problem instance $\langle P,\kappa\rangle$, $\kappa$ \emph{can, in principle, be freely chosen}.

Just as in MBD, when minimum-cardinality or most probable diagnoses are sought, there are two main factors affecting the complexity of solving a JOP instance, i.e., the efficiency of consistency checks and the cardinality of the solution(s) $\Delta^*$. 
In case consistency checking is too slow (\emph{excessive runtime}) or solution size is too large (\emph{excessive memory requirements}), the JOP instance will become practically infeasible. The two observations (i) and (ii) provide a remedy for both issues, if suboptimal solutions (wrt.\ the sum of job utilities, cf.\ Def.~\ref{def:JOP_problem}) are acceptable. 

Specifically, (i) allows to flexibilize the definition of consistency in order to allow \emph{approximate consistency checks} and thereby reduce consistency checking cost. The point is to leverage neighborhoods $[\kappa,\kappa + \varepsilon]$ to define in a natural way what "almost consistent" or "slightly inconsistent" means\footnote{Sometimes it might be more useful to define the neighborhood relative to $\kappa$. In this case, $\varepsilon$ can be defined as a function $\varepsilon_\kappa$ of $\kappa$. E.g., if a five percent deviation from $\kappa$ should be tolerated, then $\varepsilon_\kappa := 0.05\kappa$.} (note, this is, if at all, not as naturally possible for logical consistency). 
In fact, replacing "consistent" ($\cost(S) \leq \kappa$) with "almost consistent" ($\cost(S) \leq \kappa + \varepsilon$) can lead to significant performance improvements by allowing for an early termination of the schedule solver in certain cases.\footnote{The rationale behind these potential gains is that scheduling-problem solvers often find very good (sub)optimal solution candidates rather quickly and actually spend most of their computation time 
achieving only marginal solution improvements.
When an early found suboptimal candidate falls within the cost region $[\kappa,\kappa+\varepsilon]$, then substantial runtime savings can be the result.} For that reason, state-of-the-art scheduling engines, like IBM's CP-Optimizer, allow to specify such an optimality tolerance $\varepsilon$.   
Note, this $\varepsilon$-redefinition of consistency preserves the monotonicity of the consistency predicate, i.e., any subset (superset) of a consistent (inconsistent) job set is consistent (inconsistent). In addition, when using hitting-set-based solving mechanisms such as uniform-cost HS-Tree \cite{Reiter87,greiner1989correction,Rodler2015phd} or one of its derivatives \cite{rodler2020ecai,rodler2018statichs} as a JOP solving mechanism, the suboptimality (excess cost beyond $\kappa$) of the computed solution is bounded by $\varepsilon$. 
%

On the other hand, (ii) can be exploited to \emph{iteratively solve a series of JOP instances, each with suitably modified $\kappa$ to keep solutions small and preserve solvability}. The idea is to rely on a "floating" definition of consistency: Given a JOP instance $\langle P_\jobs,\kappa\rangle$ where $P_\jobs$ denotes a job scheduling problem with job set $\jobs$, first solve a relaxed JOP instance $\langle P_\jobs,\kappa_1\rangle$ for $\kappa_1 > \kappa$. Use the solution $\Delta^*_1 \subseteq \jobs$ for this problem to specify a less relaxed problem $\langle P_{\jobs\setminus\Delta^*_1},\kappa_2\rangle$ for $\kappa \leq \kappa_2 < \kappa_1$. Continue this process until $\kappa_n = \kappa$. Define the overall solution as the union $\Delta^* := \Delta^*_1 \cup \dots \cup \Delta^*_n$. Finally, verify (and if necessary, establish by deleting elements using a polynomial procedure like QuickXPlain \cite{junker04}) the subset-minimality of $\Delta^*$. In this vein, one does solve multiple, but likely easier JOP problems (given that $\kappa_1$ is sufficiently large and $|\kappa_i - \kappa_{i+1}|$ is sufficiently small).

\noindent\textbf{Background:}
    Some works in MBD \cite{Shchekotykhin2012,Rodler2013,DBLP:conf/icbo/SchekotihinRSHT18,junker04} use the concept of a background $\mb$, which encompasses system components that are known or assumed to be fault-free (e.g., because they have already been tested or approved in some other way). 
    More generally spoken, $\mb$ is a tool to \emph{focus or reduce the search space} for diagnoses in that every component in $\mb$ must not occur in any diagnosis or conflict. 
    Formally, if a background $\mb$ is given for a DPI, this implies that 
    the system description is reformulated as $\sd_\mb := \sd \cup \{\ok(c)|c\in\mb\}$ and that the predicate $\OK()$ is defined only over $\comps \setminus \mb$, i.e., no ($\ok$ or $\lnot\ok$) assumptions about elements in $\mb$ are allowed.
    %
    
    Such a mechanism also comes in handy when considering JMP/JOP. It is not far to seek practical use cases. For instance, assume a company has some key customers and several other, less important ones. In such a scenario, the company would perhaps strive to gratify its key customers in the first place, and to give the production 
    of their orders \emph{absolute} priority over other products. This can be contrived by simply adding the prioritized jobs to $\mb$, which corresponds to the constraint that all jobs in $\mb$ \emph{must} be in the schedule. Note that the same effect could be achieved by assigning non-zero utility to all prioritized jobs, and zero utility to all others. However, in this case no \emph{relative} prioritization of less important jobs is possible. Hence, using $\mb$ in combination with utilities assigned to jobs in $\jobs\setminus\mb$, both \emph{absolute and relative priorities can be specified}. 
    
    Three remarks: \emph{(1)}~Although any job in the background $\mb$ must not occur in any found solution $\Delta^*$ for JMP/JOP, all jobs in the background take part in all computations (e.g., consistency checking and conflict computation). \emph{(2)}~The approach to simply exclude background jobs $J$ from $\jobs$ does not equivalently simulate the usage of a background $\mb := J$ because (unless a concomitant adequate adaptation of $\fw$ takes place) this would imply that jobs $J$ will not be factored in when the schedule is computed. \emph{(3)}~If too large a set of jobs is added to $\mb$ such that there is no schedule $S$ for $\jobs:=\mb$ with $\cost(S) \leq \kappa$, then the JMP/JOP instance becomes unsolvable \cite{Rodler2015phd}. Hence, an initial test is necessary to verify solvability.    
    %
    %

\noindent\textbf{Conflicts:} 
    In MBD, a conflict is a set of components not all of which can be normal (as otherwise we can derive a logical contradiction). In JOP, a conflict is a set of jobs not all of which can be in the schedule (as otherwise we can derive that the schedule's cost to be larger than $\kappa$). Just as conflicts are used in MBD for analysis purposes and as intermediate results towards determining diagnoses (as hitting sets of conflicts), conflicts can serve the same purpose in JOP. 
    
    Nevertheless, one material difference 
    is that, in MBD, minimal conflicts $\mc_i$ tend to be small in size compared to the number of system components, i.e., $|\mc_i| \ll |\comps|$, whereas, in JOP, the size of minimal conflicts might often be in the scale of $|\jobs|$, i.e., $|\mc_i| \approx |\jobs|$. The latter will especially hold if the available job processing time $\kappa$ is close to the least time $\kappa^*$ required to process \emph{all} jobs in $\jobs$, i.e., $\frac{\kappa}{\kappa^* - \kappa} \gg 1$. This difference has to be borne in mind when devising algorithms for solving JOP.
    
    For example, one consequence of large-sized minimal conflicts is that 
    the set $\mathbf{C}$ of all conflicts will be large, which means that the set of all diagnoses $\mathbf{D}$ is small (as $|\mathbf{C}|+|\mathbf{D}| = 2^{|\comps|}$ \cite{Slaney2014}), which in turn implies that the size of minimal diagnoses is low. In the extreme case the minimum-cardinality diagnoses have size one. Coupled with the flexible definition of sub-goals $\kappa_i$ discussed in the "Consistency Checking" paragraph above, it might be possible to reduce the overall JOP instance to a series of instances where each can be solved by a singleton set $\Delta$.

    As to the computation of minimal conflicts for JMP/JOP, the same algorithms as used in MBD can be adopted,
    e.g., MSMP (Minimal Set subject to a Monotone Predicate) algorithms \cite{junker04,rodler2020qx,shchekotykhin2015mergexplain,marques2013minimal}.
    They perform $O(|\jobs|)$ many calls \cite{marques2013minimal} to a scheduling-problem solver for the computation of one minimal conflict.
    For instance, QuickXPlain \cite{junker04,rodler2020qx} splits the set $\jobs$ into two subsets $J_1,J_2$ and recursively analyzes $J_1$ and $J_2$ to find a minimal subset $\mc \subseteq \jobs$ such that for $P_\mc = \langle \mc,\fw,\cost\rangle$, there is no schedule $S_\mc$ with $\cost(S_\mc) \leq \kappa$.

    

\noindent\textbf{Diagnoses:} 
    In classic MBD, a diagnosis is an explanation (set of faulty components) for the discrepancy between the initial belief (properly working system) and reality (observed system misbehavior). Inherently, there is just one correct diagnosis in MBD. The direct counterpart of a minimal diagnosis in the scheduling domain is a minimal job set $\Delta$ that solves the JMP problem (cf.\ Def.~\ref{def:JOP_problem}). 
    Depending on the specified job utilities, 
    the MBD-equivalents to the JOP-solutions 
    $\Delta^*$ (cf.\ Def.~\ref{def:JOP_problem}) are the minimum-cardinality diagnoses (uniform utilities) or the most preferred diagnoses (non-uniform utilities). In general, there might be multiple solutions in JOP. In fact, 
    given the maximal production time $\kappa$, executives of a company 
    can be expected to be indifferent to 
    which jobs are eliminated from the job set, as long as there is a schedule consistent with all formulated requirements with optimal utility (e.g., revenue after the products corresponding to the finished jobs are sold).
    %

\noindent\textbf{Diagnosis Costs:}
    In MBD, diagnosis costs are often defined from the probabilistic perspective, where components $c$ have an (estimated) failure probability $p_c$ and the probability of a diagnosis (to be the correct solution) is specified \emph{in a multiplicative way} under the assumption that components fail independently.
    More precisely, the probability $p(\md)$ of a diagnosis $\md$ (assumption that components in $\md$ are abnormal, and those in $\comps\setminus\md$ are normal) is characterized as $\prod_{c \in \md} p_c \prod_{c \in \comps\setminus\md} (1-p_c)$ \cite{dekleer1987}.
    
    As opposed to this, it appears natural in JOP to define diagnosis utility \emph{in an additive way} 
    over
    job utilities (cf.\ Def.~\ref{def:JOP_problem}) since---at least in the straightforward use cases---costs, profit, revenues or the like are of interest. Obviously, the proper way of combining job utilities of this type is by summing them up, e.g., one more product sold yields the price of the product more revenue.  
    
    This difference might be important when it comes to the computation of solutions.
    For instance, while an open problem in MBD for multiplicative costs, there are ideas of heuristics for best-first diagnosis computation in the case of additive costs \cite{meilicke2011}. Such results might be worth taking into account when approaching JOP.
    

\noindent\textbf{Observations, Measurements and Oracle:} In JMP/JOP, any additional information or hint about the (unknown sought) consistent schedule with cost less than $\kappa$ can be interpreted as an observation or measurement. In MBD, system observations and measurements are mostly employed to dismiss spurious diagnoses. 
    
    \noindent\emph{How to Determine Interesting Questions:} 
    Diagnosis systems often come up automatically with useful suggestions of measurement 
    points in the diagnosed system. 
    What such a measurement recommendation could look like in JMP/JOP is not immediately clear. A \emph{conflict-based} way to approach the determination of informative measurement points (or: questions to ask about the sought schedule) is the idea to determine measurements by reasoning over partial conflicts \cite{DBLP:journals/corr/Rodler16a}. 
    Another \emph{diagnosis-based} strategy is to precompute some minimal diagnoses and to analyze their structure (common elements and entailments) in order to figure out useful questions to an oracle \cite{Shchekotykhin2012,Rodler2015phd}. More specifically, questions computed by these strategies, regardless of their answer, reduce one (or more) minimal conflict(s) in size and therefore eliminate some spurious diagnoses (set-minimal job sets $\Delta$). As the space of such questions may be large, reaching thousands of elements \cite{DBLP:journals/corr/Rodler2017}, heuristics \cite{dekleer1987,Shchekotykhin2012,Rodler2013,rodler17dx_activelearning,rodler2018ruleML} are often leveraged to find a "best" question to pose.
    
    \begin{example} \hspace{-1em}\emph{(Oracle Queries in JOP)}\quad \label{ex:oracle_queries_in_JOP}
    Let us get back to our problem instance discussed in Example~\ref{ex:job_shop_scheduling_illustration_of_JMP+JOP}. Once having deduced that $\{\text{job 1},\text{job 4}\}$ is a minimal conflict in our example, one could ask the question 
    "Should job 4 remain in the schedule?".
    If a consulted oracle negates this question, then JOP would be immediately solved. The reason is that a conflict computation for $\jobs\setminus\{\text{job 4}\}$ would return 'no conflict' (since each original minimal conflict contains $\text{job 4}$ or, equivalently, $\{\text{job 4}\}$ is a minimum-cardinality diagnosis, cf.\ Example~\ref{ex:job_shop_scheduling_illustration_of_JMP+JOP} and Fig.~\ref{fig:exAll}e). In the affirmative case, we would find that both $\text{job 1}$ and $\text{job 3}$ must not be in the schedule (since $\{\text{job 1},\text{job 4}\}$ and $\{\text{job 3},\text{job 4}\}$ were originally minimal conflicts, which yields new conflicts $\{\text{job 1}\}$ and $\{\text{job 3}\}$ since $\text{job 4}$ must be in the schedule). Eventually, a conflict check for $\jobs\setminus\{\text{job 1},\text{job 3}\}$ would output 'no conflict', thereby signalizing that the minimum-cardinality diagnosis in this case is $\{\text{job 1},\text{job 3}\}$.
    \end{example}
    
    \noindent\emph{Goal of Oracle Interactions:}
    In MBD, interaction with an oracle helps to discriminate between competing solutions and to gain information about the \emph{(single) correct solution}; it is a means to \emph{learn more about reality}. Without information acquisition beyond the original DPI, a diagnosis system is simply unable to differentiate between a set of diagnoses---all of them satisfy the definition of a solution. However, if the DPI is extended with measurements, the DPI changes, and so does the set of solutions.
    
    In JOP, by contrast---assuming that all requirements regarding the desired schedule are stated in $\fw$ and all jobs that have to necessarily remain in the schedule have been shifted to the background $\mb$---oracle inquiries serve a crucially different purpose. 
    Because \emph{(i)}~there might be \emph{multiple solutions} (with equal utility), \emph{(ii)}~there is no notion of "incorrect", but rather of "non-optimal", for solution candidates, and \emph{(iii)}~even approximations of solutions (with slightly suboptimal utility) might suffice, oracle consultations in case of JOP can be seen as an instrument to \emph{make the solution process more efficient}. 
    
    \noindent\emph{Who Acts as an Oracle?} Depending on the type of diagnosed system, there are various natural oracles in MBD to accomplish the task of conducting measurements or making observations, e.g., an engineer in the case of a physical device or circuit, or a domain expert when debugging a knowledge base. 
    %
    Because the correctness of information given by the oracle is existential for diagnostic success in MBD \cite{rodler2019KBS_userstudy}, one usually assumes that an oracle has full (or at least partial) competence wrt.\ the diagnosed system \cite{DBLP:conf/foiks/SchekotihinRS18}. In the JOP case, on the other hand, one can somewhat less strictly entitle any procedure/entity as an oracle that we expect to be 
    sufficiently informed to reasonably guide the solution finding process. Even automatic oracles or ones based on heuristics are thinkable. For instance, if all jobs have the same utility (i.e., minimum-cardinality diagnoses are the solutions to JOP), one such heuristic an oracle could opt for is incrementally deciding to eliminate from $\jobs$ the one job that appears in most minimal conflicts (cf.\ the \emph{eminc}-procedure suggested by \cite{dekleer2011hitting}). 

\section{Related Work}
\label{sec:related_work}

As scheduling problems can be represented quite naturally as constraint satisfaction problems (CSPs), there exists a quite long tradition in the application of CSP approaches on scheduling problems. There is also some literature targeting the minimization of the number of violated constraints of over-constrained CSPs. Here, especially the approaches based on (weighted) MaxCSP (e.g., \cite{Schiex,Bistarelli,Domshlak}) are worth mentioning. Differences between (Max)CSP approaches and our approach to JMP/JOP are: 
\begin{itemize}[noitemsep]
    \item We view the set of constraints (cf.\ $\fw$ in Def.~\ref{def:job_scheduling_problem}) as a black box and operate on jobs, i.e., constraints are (unmodifiable) hard constraints, whereas (weighted) MaxCSP methods aim at finding the (most preferred) subset of constraints including a maximal number of satisfied constraints. Simply put in MBD terms, in our case jobs are the system components, while in MaxCSP constraints are the system components. One benefit of regarding the constraints as a black box is that our approach is generic in the sense that it can be applied to any scheduling problem instance, regardless of how it is modeled, and without needing to modify or adapt the model. Moreover, MaxCSP is not (at least not as naturally) amenable to the exploitation of concepts such as approximate consistency checking as JMP/JOP (cf.\ Sec.~\ref{sec:JOP_vs_MBD}).
    \item We view the problem from an MBD, and not from a CSP perspective, which involves some advantages. For instance, it allows us to easily express the subset-minimality optimization criterion, as addressed by JMP, which \emph{(i)}~is a problem of practical relevance, \emph{(ii)}~is genuinely easier (cf.\ Theorem \ref{thm1}) than MaxCSP (which is NP-hard), and \emph{(iii)}~is non-trivial to specify in terms of CSPs.
\end{itemize}

\cite{lim}, similarly as done in our work, use a "two-level" approach where a high-level component is coupled with constraint solving techniques. However, in our case, this high-level component is realized by MBD strategies, whereas \cite{lim} draw on \emph{approximate} heuristic methods. 
Moreover, they address a different optimization criterion as they do not attempt to optimize the sum of utilities of non-removed jobs, but try to maximize the number of remaining jobs of high priority based on a lexicographic order. That is, they would prefer three jobs with utilities 3,2,1 (sum is 6) to seven ones with 3,1,1,1,1,1,1 (sum is 9). 

\section{Evaluation}
\label{sec:eval}
\noindent\textbf{Goal.} This evaluation aims at getting a first impression of whether MBD offers effective tools for tackling JMP/JOP problems in terms of solution quality as well as runtime performance. To this end, we apply MBD to approach JMP instances over a well-known suite of 
scheduling benchmark problems. Based on the obtained results, we discuss potential strategies to attack the JOP problem.

\noindent\textbf{Dataset.} For the creation of the JMP evaluation dataset, we used a subset of the Taillard benchmark problems \cite{taillard} as a basis. This dataset has been heavily used in scheduling research for more than 25 years as different instance characteristics reflect different industrial realities. In particular, we used the ten largest instances comprising 100 jobs to be scheduled on 20 machines (= 2000 operations in total, i.e., one operation on each machine per job) as well as ten of the somewhat smaller instances comprising 50 jobs to be scheduled on 15 machines. One big advantage of the used benchmark problems is that optimal completion times (i.e.\ the timespan from the start of the first operation to the termination of the last operation) are known, which allowed us to systematically control the hardness of the derived JMP instances. Given a Taillard problem instance $P$ and the corresponding optimal completion time $\kappa^*$, a JMP instance can be created setting $\kappa = r \cdot \kappa^*$ with $r<1$. Clearly, the smaller $r$ is (and consequently the smaller $\kappa$ is) the more jobs must be excluded in order to produce a consistent schedule with a completion time $\leq \kappa$. To explore various scenarios that might arise for a company in case production deadlines prevent it to cover all orders, we specified five deadline levels $r \in \{0.95,0.9,0.85,0.8,0.75\}$. Thus, our test dataset consists of 20 (Taillard instances) $\times$ 5 (deadline levels $r$) = 100 JMP problem instances.\footnote{Dataset available at: isbi.aau.at/ontodebug/evaluation}

\noindent\textbf{Experimental Setting.}
To perform consistency checks, i.e., tests whether there is a consistent schedule $S$ such that $\cost(S) \leq \kappa$, 
we used IBM's CP-Optimizer\footnote{www.ibm.com/analytics/cplex-cp-optimizer} which is currently one of the most powerful scheduling optimization tools \cite{cspsched}.
To compute diagnoses, i.e., solutions to the JMP problem, 
we used the \emph{Inverse QuickXPlain} algorithm (Inv-QX \cite{Shchekotykhin2014}) which is a modified version of the \emph{QuickXplain} algorithm \cite{junker04,rodler2020qx}.
In general, Inv-QX takes a diagnosis problem instance (JMP/JOP instance) $P$ and a consistency checker (schedule optimizer) as input and executes a worst-case linear (in $|\jobs|$) number of consistency checks to compute a diagnosis (solution to JMP\footnote{Although Inv-QX can be used as a heuristic method to approximate optimal diagnoses (JOP solutions) in that its input collection of components (jobs) is suitably sorted \cite{DBLP:journals/corr/Rodler2017}, no (provable) guarantees on the output quality can be given in general.}) for $P$. 

\noindent\textbf{Results.} Table~\ref{tab:results} shows the runtimes and diagnosis sizes (number of jobs eliminated from the job set) for all 100 test cases. From the table, we learn that a JMP solution for 
\begin{itemize}[noitemsep]
    \item all (100,20) instances with all\\ $r \in \{0.95,0.9,0.85,0.8,0.75\}$
    \item all (50,15) instances with $r\in \{0.95,0.9,0.85\}$
    \item 90\,\% of the (50,15) instances with $r=0.8$
    \item 60\,\% of the (50,15) instances with $r=0.75$
\end{itemize}
could be computed within a six-hour timeout period (reflects overnight calculation). For $r \in \{0.95,0.9,0.85$, $0.8,0.75\}$, the median runtimes (in sec) were $\{118,165,142,159,366\}$ for the (50,15) and $\{304,397,306,444,490\}$ for the (100, 20) instances. Maximal runtimes (for instances solved before the timeout) amounted to over 1h30min for (50,15), but to no more than 40min for (100,20). Hence, the (50,15) instances turned out to be the harder challenges in the JMP scenario. 
A closer analysis of the hardest (50,15) cases---those that exceeded the timeout---revealed that certain subsets of the 50 jobs were extremely hard-to-decide "hotspots" \cite{Goncalves2012}, i.e., the schedule optimizer required up to more than 
$10^5$sec (!) for a single consistency check. The chance of encountering such hotspots rises significantly when comparing schedule optimization with JMP/JOP since, to solve the latter, multiple calls of the schedule optimizer, each for a different subset of all jobs, are required. Note, over the rest of the (more benign) instances, an average consistency check still took a single digit number of minutes, which is by orders of magnitude more than in (hard) usual MBD scenarios, cf., e.g., \cite{Shchekotykhin2014}. This underscores the complexity of JMP/JOP. 
Still, the results overall show that MBD does indeed offer suitable out-of-the-box tools to solve many JMP 
instances of practically relevant size within reasonable time bounds.\footnote{
According to 
former project partners from the semiconductor industry, it will 
often suffice for JMP/JOP to grant "overnight timeliness", i.e., when the result is known the next day.}

\noindent\textbf{Towards JOP.}
Observe in Table~\ref{tab:results} that all found solutions for one and the same deadline level $r$ have similar size, although we used Inv-QX with an arbitrary sorting of its input, which leads to the finding of just \emph{any} solution (diagnosis). This appears to empirically confirm our intuition that most of the solutions 
for a given deadline level $r$ should have a size of approximately $|\jobs|*(1-r)$. Therefore, reducing the allowed completion time $\kappa$ for the schedule by a fraction $(1 - r)$ requires the elimination of about the same fraction of jobs from the job set to comply with the reduced deadline. Moreover, this suggests that the minimum-cardinality solutions (i.e., the JOP solutions for uniform job utilities) also gather around this size. Hence, 
\begin{itemize}[noitemsep]
    \item JMP solutions seem to constitute acceptable approximations of (uniform-job-utility) JOP solutions,
    \item a simple heuristic algorithm to approach JOP is to use Inv-HS-Tree (i.e., multiple calls to Inv-QX with a systematic modification of Inv-QX's inputs such that one and the same solution cannot be found twice) which can be expected to quickly find JMP solutions that are close to JOP ones, and
    \item this educated guess of the (minimum-cardinality) solution size could be leveraged as a key input to algorithms targeting the JOP, e.g., a depth-limited \cite{russellnorvig2010} variant of HS-Tree where $|\jobs|*(1-r)$ is used as a limit.
\end{itemize}


\begin{table}[t]
\center
\scriptsize
\begin{tabular}{c||c|c||c|c}
&\multicolumn{2}{c||}{(\textbf{50 jobs, 15 machines)}}&\multicolumn{2}{c}{(\textbf{100 jobs, 20 machines)}}\\
\textbf{r} & \textbf{diag size} & \textbf{time} & \textbf{diag size} & \textbf{time}\\
\hline
0.95&3&137&4&407\\
0.95&3&169&4&152\\
0.95&4&99&4&446\\
0.95&2&93&5&230\\
0.95&2&302&4&842\\
0.95&3&97&6&287\\
0.95&3&22&5&321\\
0.95&3&27&5&230\\
0.95&3&185&4&200\\
0.95&3&207&5&333\\
\hline
0.9&7&91&10&392\\
0.9&5&250&10&209\\
0.9&6&223&9&965\\
0.9&5&42&10&401\\
0.9&5&425&9&803\\
0.9&6&36&11&166\\
0.9&6&34&10&391\\
0.9&5&107&10&375\\
0.9&5&375&9&409\\
0.9&5&453&9&834\\
\hline
0.85&9&42&15&358\\
0.85&8&651&18&264\\
0.85&9&664&14&1092\\
0.85&8&30&15&313\\
0.85&7&657&14&991\\
0.85&10&34&17&173\\
0.85&7&178&15&476\\
0.85&8&106&15&217\\
0.85&7&5250&15&230\\
0.85&8&68&15&299\\
\hline
0.8&13&28&20&490\\
0.8&11&571&24&654\\
0.8&---&---&21&759\\
0.8&10&99&21&266\\
0.8&10&450&20&790\\
0.8&11&42&21&344\\
0.8&11&159&19&397\\
0.8&10&613&19&688\\
0.8&11&3093&20&291\\
0.8&12&140&20&350\\
\hline
0.75&13&79&25&956\\
0.75&14&5497&30&446\\
0.75&---&---&25&2168\\
0.75&---&---&28&308\\
0.75&---&---&25&1830\\
0.75&13&125&28&350\\
0.75&14&66&24&943\\
0.75&13&607&25&389\\
0.75&---&---&24&534\\
0.75&14&695&26&434\\
\end{tabular}
\caption{Results for JMP. "---" indicates timeout (6h).}
\label{tab:results}
\end{table}

\section{Conclusions and Future Work}
\label{sec:conclusion}

In this work, we motivate and formally define the job-set maximization (JMP) and the job-set optimization (JOP) problems for scheduling. Both problems address the issue which product-related jobs 
to eliminate from a schedule in case not all jobs can be accomplished with available (time) resources. 
To gain an understanding of the idiosyncracies of JMP/JOP and how to deal with them from a model-based diagnosis perspective, we thoroughly discuss similarities and dissimilarities between JMP/JOP and conventional diagnosis problems. Finally, a proof-of-concept evaluation shows that existing model-based diagnosis techniques can be successfully applied on a benchmark of industrial-scale scheduling problems.

Future work topics include:
\begin{itemize}[noitemsep]
    \item Development, study and evaluation of 
    algorithms 
    for tackling 
    JOP based on 
    discussions in Sec.~ \ref{sec:JOP_vs_MBD} and \ref{sec:eval}.
    \item Investigation of JMP/JOP on further manifestations of the job scheduling problem (see Def.~\ref{def:job_scheduling_problem}), such as open shop or flow shop problems \cite{Blazewicz:2007}.
    \item Deeper analysis of the JMP instances that could not be solved within the allowed time limits.
    \item Tuning of the scheduling engine in order to improve consistency checking performance, which turned out to have a major impact on runtime in our experiments.
\end{itemize}

\section*{Acknowledgments}
This work was supported by the Austrian Science Fund (FWF), contract \mbox{P-32445-N38}. Thanks to Dietmar Jannach for valuable comments that helped improve this manuscript. 

\fontsize { 9pt }{ 9pt } 
\selectfont

\balance

\end{document}